\documentclass[aps,letterpaper,english,reprint,showkeys,floatfix]{revtex4-1}
\usepackage[T1]{fontenc}
\usepackage[latin9]{inputenc}
\usepackage{babel}
\usepackage{calc}
\usepackage{amsmath}
\usepackage{amssymb}
\usepackage{graphicx}
\usepackage{wrapfig}
\usepackage{float}
\usepackage{cleveref}
\usepackage{multirow}
\usepackage{setspace}
\usepackage{verbatim}
\usepackage[bottom=0.8in,top=0.8in,right=.5in,left=.75in]{geometry}
\setcitestyle{authoryear}
\makeatletter
\usepackage{parskip} 

\pdfpageheight\paperheight
\pdfpagewidth\paperwidth

\providecommand{\LyX}{\texorpdfstring%
  {L\kern-.1667em\lower.25em\hbox{Y}\kern-.125emX\@}
  {LyX}}
\makeatother

\begin{document}

\title{Recombinant dynamical systems}
\author{Saul Kato}
\email{corresponding author: saul.kato@ucsf.edu}

\selectlanguage{english}%

\affiliation{ Weill Institute for Neurosciences, Department of Neurology\\ University of California, San Francisco}

\date{\today}

\begin{abstract}
%\linenumbers
\noindent We describe a connectionist model that attempts to capture a notion of experience-based problem solving or task learning, whereby solutions to newly encountered problems are composed from remembered solutions to prior problems. We apply this model to the computational problem of \emph{efficient sequence generation}, a problem for which there is no obvious gradient descent procedure, and for which not all posable problem instances are solvable. Empirical tests show promising evidence of utility.
\end{abstract}

\maketitle

%============================================================================================

\section{\label{sec:level1}Introduction}
%\linenumbers

 How do we solve problems based on past experience? We posit that remembered solutions to solved problems exist as persistent constructs in the brain of the solver. To be recalled at a later date, a solution must be played back in time; therefore, such a persistent construct may be viewed as a specification of a dynamical system. If a memorized solution is to be accessible at will for the problem solver, it must be evaluable, i.e. ``runnable'' again in the mind-space of the solver. Do these replayable constructs exist separably from other similar constructs? If so, can they be somehow brought together, or re-combined, to solve new problems? Despite a lack of evidence from experimental neuroscience of such recombinable subunits -- and an absence of such explicit architectures in the current state-of-the-art large machine intelligence models -- we hypothesize that some of our powerful and mysterious human cognitive abilities rely on the compositionality of remembered, separable, successful solutions.

We define a computational model that captures the idea of recombinability of dynamical systems and apply it to a rudimentary computational problem which we term efficient sequence generation. Efficient sequence generation bears analogy to Turing's question of computable numbers. The model has obvious relation to finite state automata and evolutionary algorithms.

%============================================================================================
\section{\label{sec:level1}Definitions}

\underline{Definition}: A Boolean network machine (\textbf{BNM}) is a quadruple $(X,E,I,o)$ where
\begin{itemize}
    \item $X$ is a set of boolean function nodes
    \item $E$ is a set of directed edges, $E \subseteq X \times X$
    \item $I$ is a set of input nodes, $I \subseteq X$, possibly $\varnothing$ 
    \item $o$ is a single element of X, called the output node
\end{itemize}

A BNM keeps a binary state on each node and thus defines a dynamical system. We initialize all nodes to zero. A finite BNM will always eventually enter a state cycle of at most $2^N$ states, since the system is deterministic and there are only $2^N$ possible states.

\underline{Definition}: A \textbf{bag} is a set of BNMs.

\underline{Definition}: A \textbf{gluing rule} is a well-defined, possibly randomized, procedure that takes two BNMs and makes a single BNM out of them.

\begin{figure}[H]
    \centering
    \includegraphics[width = 0.44\textwidth]{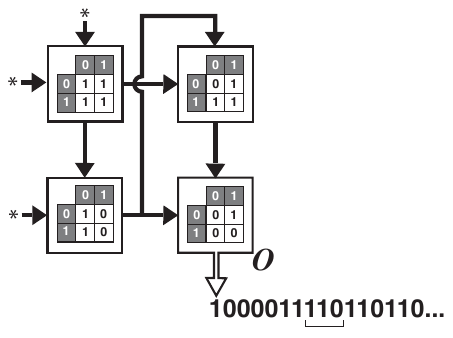}
    \caption{A ZISO-2BNM (or simply 2BNM here) of size 4 with output $\textbf{110}$. Each node's boolean function is represented as a truth table. Asterisks denote irrelevant inputs.}
\end{figure}

   \begin{figure}[H]
        \includegraphics[width = 0.44\textwidth]{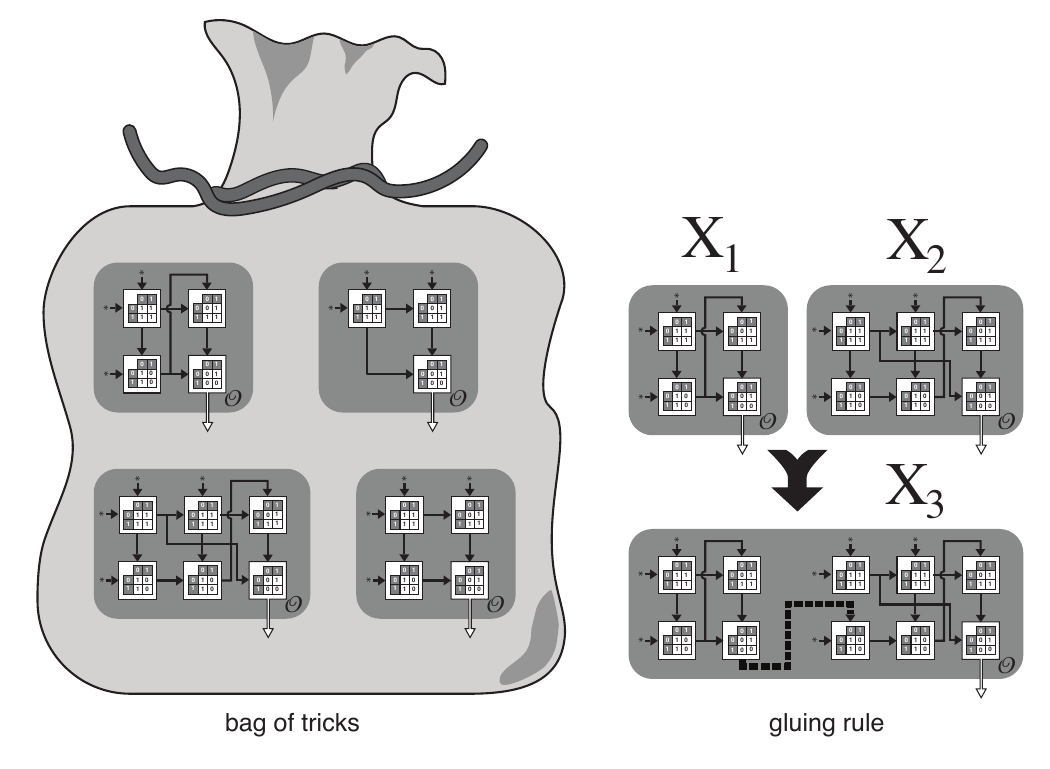}
    \caption{A bag of BNMs and a gluing rule: glue the output of $X_1$ to one randomly selected input of a node in $X_2$} 
\end{figure}

\underline{Further definitions}:

\begin{itemize}

\item a \textbf{c-string} is the set of a string $\{0,1\}^{*}$ and all of its rotations.

\item the \textbf{state} of a BNM is a list of symbols, one for each node, representing the instantaneous dynamical state

\item the \textbf{output} of a BNM is the c-string produced on the output node after the network has reached a state cycle. We always initialize all nodes to zero.

\item the \textbf{size} of a BNM is the number of nodes $N=|X|$.

\item a \textbf{kBNM} is a BNM that consists solely of nodes of k-ary functions, i.e. with k inputs. A kBNM is equivalent to a Kaufmann NK network with a set of designated input nodes and 1 output node \citep{kauffman1993origins}. A zero-input single output (\textbf{ZISO}) BNM has no input nodes. \textbf{We only study ZISO-2BNMs in this paper, hereafter denoted simply as \textbf{2BNMs}.}
    
\item A BNM $X$ is \textbf{efficient} with respect to a c-string $s$ if $|X| << |s|$, or to be precise, it is $k$-efficient if $|X| < \log_2(|s|)-k $.
    
\item a node $n_i$ is \textbf{hidden} if $n_i \in X - (I \cup O)$, i.e. it is an internal node that only communicates with other nodes.
    
\end{itemize}

\section{\label{sec:level2}The problem}

Here's the challenge: can we find 
efficient 2BNMs? That is to say, is there a tractable (i.e. space and time bounded) procedure for finding efficient 2BNMs in the space of all possible 2BNMs? More rigorously we define ESG as:

Given an integer s, can we define a SPACE(s) and TIME(s) bounded procedure to find an O(log(s))-sized set of efficient 2BNMs (each of max size O(log(s)) that generates c-strings of length s)? We call this problem \textbf{ESG}.

Here are two intuitive propositions regarding \textbf{ESG}:

PROPOSITION: not all strings have efficient BNMs.
ARGUMENT: Not all strings are compressible; in fact, most are not \citep{chaitin1987algorithmic}. A BNM of size N can be implemented on a universal computer in SPACE(N) and TIME(N). If a string had an efficient BNM, this would be a form of compression. Therefore, an incompressible string cannot be the output of an efficient BNM.

PROPOSITION: If an efficient BNM with inputs exists for a c-string s, then an efficient BNM with no input nodes exists that outputs that c-string.

ARGUMENT: There is a trivial construction of a BNM where inputs are replaced by nodes with constant truth table outputs.

\section{\label{sec:level2}Empirical evidence for \textbf{ESG} being a hard problem}

We perform some empirical studies of 2BNMs to rule out trivial solutions to \textbf{ESG}.

I.  We can't easily guess at efficient 2BNMs. Output lengths of randomly chosen 2BNMs appear to follow a power-law-like distribution (Fig. 3). Our random procedure is to select $X$  and $E$ for each node from independent uniform random distributions.

\begin{figure}[H]
        \includegraphics[width = 0.44\textwidth]{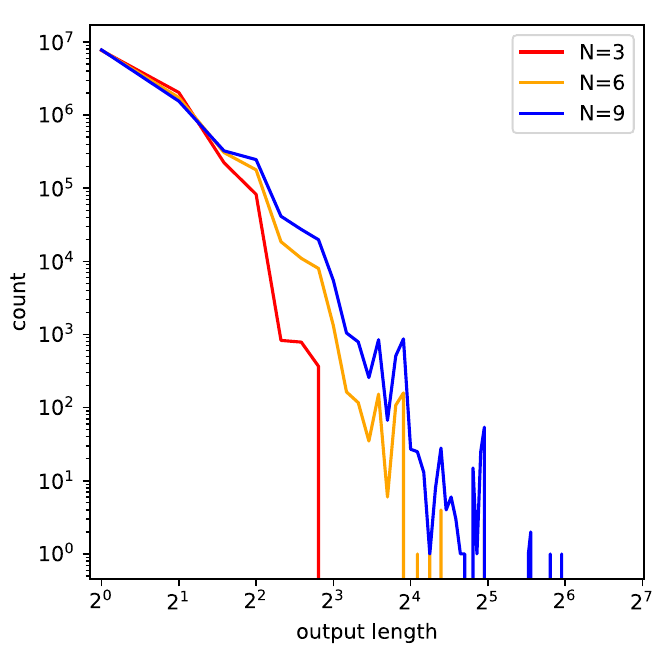}
    \caption{Output lengths of 10 million randomly chosen 2BNMs of size 3, 6, and 9 appear to follow a power law.}
\end{figure}

 This observation seems analogous to the observation that the lengths of the state cycles of random $(K=2)$ NK networks appear to follow a power-law-like distribution \citep{kauffman1993origins}.
By contrast, lengths of state cycles of $K>=3$ NK networks approach $2^N$.

II. We cannot obviously employ local search or hill climbing/gradient ascent to find efficient BNMs. This is supported by two empirical observations (not plotted here): (1) the local landscape around good solutions (efficient BNMs) is not smooth, and (2) hill climbing from random initial choices of BNMs does not yield efficient BNMs.

If we cannot easily guess at solutions, and we cannot use gradient descent, what other problem solving schemes could we try? We ask: can we use a memory, i.e. can we exploit the knowledge of other smaller efficient 2BNMs? We imagine a human carrying around bag of tricks they acquire over their lifetime, and bring to bear on new problems. How can these tricks be exploited? We imagine a crude, unconscious brainstorming procedure, where putative new solutions are attempted simply by squashing previously memorized solutions to other problems together, which we simulate in the next section.

\section{\label{sec:level2}THE RECOMBINATION PROCEDURE}

We start with a bag of just a few efficient BNMs (found, say, by random search) and perform the following procedure:
\begin{enumerate}
\item randomly pick two BNMs $X_1$ and $X_2$ from the bag
\item glue them together into a large BNM $X_3$
\item evaluate the output of $X_3$
\item save $X_3$ to our bag if good (according to a performance metric)
\item repeat
\end{enumerate}

We pick a very simple gluing rule: attach the output of $X_1$ to a randomly chosen node input from $X_2$.

We find that this procedure is better than random search for finding new efficient 2BNMs (Fig. 4).

\begin{figure}[H]
        \includegraphics[width = 0.44\textwidth]{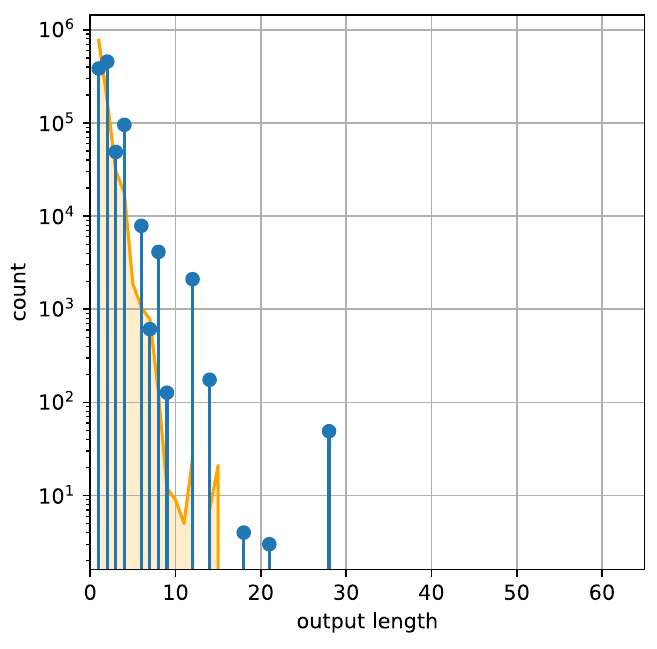}
    \caption{In blue, the distribution of output lengths of 1 million randomly generated 2BNMs of size 6 by gluing randomly selected 2BNMs of size 3 with output length 6 and 7. For comparison, the distribution of output lengths of 1 million randomly generated 2BNMs of size 6 is shown in orange.}
\end{figure}

\section{\label{sec:level2}DISCUSSION}

By storing and recombining solutions to smaller problems, our procedure seems to find efficient 2BNMs in polynomial time, in contrast to a random search. And we haven't been able to come up with a hill climbing procedure, say by choosing an appropriate representation of a BNM specification, that finds solutions.

It is intriguing that our naive recombine-and-test procedure yields a method to discover efficient 2BNMs. However, there are some gotchas that come to mind that might render the observed phenomenon trivial:

\begin{itemize}
\item Our empirical tests are deceiving and as $N$ increases, our trick will fail to work vanishingly frequently.

\item We have underestimated the space and time requirements of our process. If either blows up combinatorially, we are no better off than random search or simple enumerative strategies.

\item  We have overlooked a simple reason why this recombination procedure works.

\item  The efficient c-strings discoverable by this procedure are categorizable in a simple way. This would limit the generality of our result.

\end{itemize}

To get really excited about further exploring and developing this model, we would like to find a more interesting problem class that we can use our recombination trick for, for example:

\begin{itemize}
\item compressing a specified string

\item finding BNMs that can learn arbitrary or desirable input/output functions, both static or dynamic

\item finding BNMs that perform useful extrapolation

\end{itemize}

\section{\label{sec:level2}Code}

Python code: github.com/focolab/recombinant-dynamics

%\linenumbers

%\nolinenumbers

%\printbibliography %for biblatex
%\section*{References}
%\bibliographystyle{abbrvnat}
\setcitestyle{numbers}
\bibliography{biblio}

\end{document}